\pdfoutput=1

\documentclass[11pt]{article}

\PassOptionsToPackage{dvipsnames,usename}{xcolor}

\usepackage[final]{assets/acl}

\usepackage{adjustbox}
\usepackage{multirow}
\usepackage{wrapfig}
\usepackage{scalerel}
\usepackage{float}
\usepackage[caption = false]{subfig}

\usepackage[T1]{fontenc}    

\usepackage[utf8]{inputenc}

\usepackage{microtype}
\usepackage{svg}

\usepackage{inconsolata}

\usepackage[normalem]{ulem} 
\usepackage{twemojis}
\usepackage{soul}
\usepackage{times}
\usepackage{latexsym}
\usepackage{datetime}       
\usepackage{booktabs}       
\usepackage{nicefrac}       
\usepackage{xcolor}         
\usepackage{url}            
\usepackage{fnpct}           
\usepackage{enumitem}        
\usepackage{tcolorbox}       
\usepackage{csquotes}        
\usepackage{xspace}          
\usepackage{graphicx}

\makeatletter
\xspaceaddexceptions{\csq@qclose@i}
\makeatletter

\usepackage[scaled=0.85]{helvet}

\usepackage{bbm}
\usepackage{amsfonts}
\usepackage{amssymb}
\usepackage{mathtools}
\usepackage{amsmath}
\usepackage{bm}
\usepackage{amsthm}
\usepackage{thmtools} 
\usepackage{thm-restate}
\usepackage{cancel}

\usepackage{nicefrac}

\usepackage[createShortEnv,commandRef=Cref]{proof-at-the-end}

\newtheoremstyle{resultstyle} 
  {.8em} 
  {.8em} 
  {} 
  {} 
  {\bfseries} 
  {.} 
  {.5em} 
  {\thmname{#1 }\thmnumber{#2. }\textbf{\thmnote{#3}}}

\theoremstyle{resultstyle}
\newtheorem{result}{Result}

\usepackage{algorithm}
\usepackage{algpseudocode}  
\makeatletter
\algnewcommand{\LineComment}[1]{\Statex \hskip \ALG@thistlm \textcolor{blue}{// #1}}
\algnewcommand{\FirstLineComment}[1]{\Statex \hskip\ALG@tlm \textcolor{blue}{\(\triangleright\) #1}}
\algnewcommand{\InlineComment}[1]{\hfill\textcolor{blue}{\(\triangleright\) #1}}
\makeatother

\usepackage{cleveref}
\crefname{section}{\S}{\S\S}
\Crefname{section}{\S}{\S\S}
\crefformat{section}{\S#2#1#3}
\crefname{figure}{Fig.}{Fig.}
\crefname{alg}{Alg.}{Alg.}
\crefname{line}{line}{lines}
\crefname{appendix}{App.}{App.}
\crefname{equation}{eq.}{eqs.}
\crefname{table}{Table}{Tables}
\crefname{proposition}{Proposition}{Propositions}
\crefname{assumption}{Assump.}{Assumps.}
\crefname{lemma}{Lemma}{Lemmas}
\crefname{definition}{Defn.}{Defns.}
\crefname{hypothesis}{Hypothesis}{Hypotheses}
\crefname{estimator}{Estimator}{Estimators}
\crefname{theorem}{Theorem}{Theorems}
\crefname{thm}{Theorem}{Theorems}
\crefname{result}{Result}{Results}

\usepackage{tabularray}
\usepackage{cellspace}
\newcommand\cincludegraphics[2][]{\raisebox{-0.3\height}{\includegraphics[#1]{#2}}}
\usepackage{setspace}

\usepackage{siunitx}
\newcommand{\q}[2]{\qty[mode=math]{#1}{#2}\xspace}

\DeclareSIUnit[quantity-product = {}, reset-math-version = false]\thousand{k}
\DeclareSIUnit[quantity-product = {}, reset-math-version = false]\million{M}
\DeclareSIUnit[quantity-product = {}, reset-math-version = false]\billion{B}
\DeclareSIUnit[quantity-product = {}, reset-math-version = false]\trillion{T}

\DeclareSIUnit[quantity-product = {}, reset-math-version = false]\x{x}
\DeclareSIUnit[quantity-product = {}, reset-math-version = false]\percent{\%}

\DeclareSIUnit[quantity-product = {}, reset-math-version = false]\hour{h}
\DeclareSIUnit[quantity-product = {}, reset-math-version = false]\min{m}
\DeclareSIUnit[quantity-product = {}, reset-math-version = false]\sec{s}

\newcommand{\integer}[1]{%
    \num[
        mode = math,
        round-mode=places,
        round-precision=0,
        group-separator={,},
        group-minimum-digits=4,
        ]{#1}%
    \xspace
}

\newcommand{\float}[2][1]{%
    \num[
        group-digits=false, 
        round-precision=#1, 
        round-mode=places,
    ]{#2}%
    \xspace}

\newcommand{\snum}[1]{\num{#1}\xspace}

\xspaceaddexceptions{\textsubscript}

\makeatletter
\DeclareRobustCommand*{\escapeus}[1]{%
  \begingroup\@activeus\scantokens{#1 }\endgroup}
\begingroup\lccode`\~=`\_\relax
   \lowercase{\endgroup\def\@activeus{\catcode`\_=\active \let~\_}}
\makeatother

\newcommand{\myemph}[1]{\textsf{{\escapeus{#1}}}}

\usepackage[textsize=tiny]{todonotes}

\newcommand{\defn}[1]{\textbf{#1}}

\definecolor{70M}{HTML}{003049}
\definecolor{160M}{HTML}{2a9d8f}
\definecolor{410M}{HTML}{d62828}
\definecolor{1.4B}{HTML}{9b5de5}
\definecolor{2.8B}{HTML}{ff9f1c}
\newcommand{\sevenmil}{\myemph{\textcolor{70M}{\q{70}{\million}}}\xspace}
\newcommand{\sixmil}{\myemph{\textcolor{160M}{\q{160}{\million}}}\xspace}
\newcommand{\fourmil}{\myemph{\textcolor{410M}{\q{410}{\million}}}\xspace}
\newcommand{\onebil}{\myemph{\textcolor{1.4B}{\q{1.4}{\billion}}}\xspace}
\newcommand{\twobil}{\myemph{\textcolor{2.8B}{\q{2.8}{\billion}}}\xspace}

\newcommand{\smalldots}{...}

\newcommand{\norm}[1]{\left\lVert#1\right\rVert}
\newcommand{\per}{\mathrm{PER}}
\newcommand{\er}{\mathrm{ER}}
\newcommand{\cka}{\mathrm{CKA}\xspace}

\newcommand{\token}{t}
\newcommand{\sequence}{\boldsymbol{\mathrm{\token}}}

\newcommand{\xdata}{\boldsymbol{x}}
\newcommand{\act}{\boldsymbol{a}}
\newcommand{\actatt}{\act^{\mathtt{ATT}}}
\newcommand{\actmlp}{\act^{\mathtt{MLP}}}
\newcommand{\attention}{\mathrm{Attention}}
\newcommand{\mlp}{\mathrm{MLP}}
\newcommand{\vtheta}{\boldsymbol{\theta}}
\newcommand{\attweight}{\vtheta^{\mathtt{ATT}}}
\newcommand{\mlpweight}{\vtheta^{\mathtt{MLP}}}
\newcommand{\actcenter}{\overline{\act}}

\newcommand{\layer}{l}
\newcommand{\numlayers}{L}
\newcommand{\seqlen}{T}
\newcommand{\residualdim}{D}
\newcommand{\hiddendim}{H}
\newcommand{\checkpoint}{c}
\newcommand{\numcheckpoints}{C}

\newcommand{\R}{\mathbb{R}}

\title{Tending Towards Stability \texttwemoji{chart increasing}:\\ Convergence Challenges in Small Language Models}

\newcommand{\camemailadress}[1]{\href{mailto:#1@cam.ac.uk}{\myemph{#1}}}

\author{
   {Richard Diehl Martinez}  ~~~~
   {Pietro Lesci}  ~~~~
   {Paula Buttery} \\
    University of Cambridge \\ \myemph{\{}\camemailadress{rd654}, \camemailadress{pl487}, \camemailadress{pjb48}\myemph{\}@cam.ac.uk}
}

\begin{document}

\maketitle

\begin{abstract}
Increasing the number of parameters in language models is a common strategy to enhance their performance. However, smaller language models remain valuable due to their lower operational costs. Despite their advantages, smaller models frequently underperform compared to their larger counterparts, even when provided with equivalent data and computational resources. 
Specifically, their performance tends to degrade in the late pretraining phase. This is anecdotally attributed to their reduced representational capacity. Yet, the exact causes of this performance degradation remain unclear. We use the Pythia model suite to analyse the training dynamics that underlie this phenomenon. Across different model sizes, we investigate the convergence of the $\attention$ and $\mlp$ activations to their final state and examine how the effective rank of their parameters influences this process. We find that nearly all layers in larger models stabilise early in training---within the first 20\%---whereas layers in smaller models exhibit slower and less stable convergence, especially when their parameters have lower effective rank. 
By linking the convergence of layers' activations to their parameters' effective rank, our analyses can guide future work to address inefficiencies in the learning dynamics of small models. 
\end{abstract}

\begin{tblr}{colspec = {Q[c,m]|X[l,m]}, stretch = 0}
    \cincludegraphics[width=1.2em, keepaspectratio]{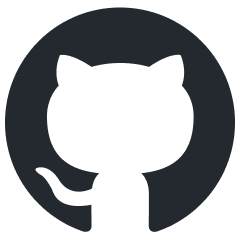} & {{\href{https://github.com/rdiehlmartinez/pretraining-playground/tree/main}{\fontsize{10.5}{12}\myemph{rdiehlmartinez/pretraining-playground}}}}
\end{tblr}

\section{Introduction}\label{sec:intro}

\begin{figure}[!t]
    \centering
    \includegraphics[width=\columnwidth]{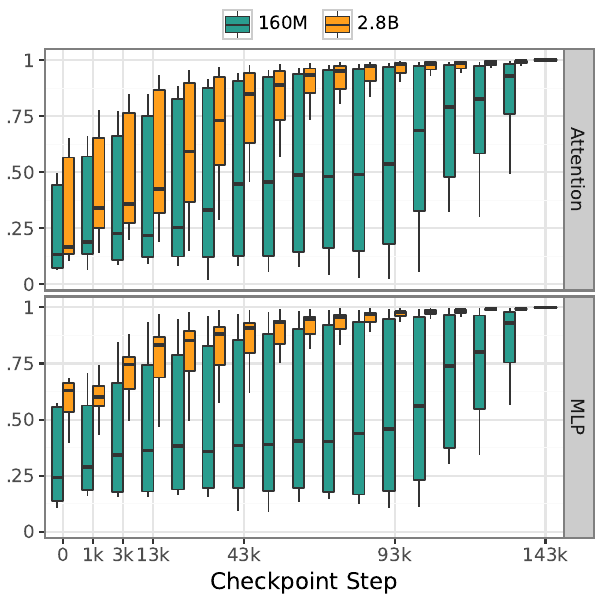}
    \caption{$\cka$ similarity (current vs.\ last checkpoint) of $\attention$ and $\mlp$ activations for Pythia \sixmil and \twobil. Distribution across layers: \integer{10}, \integer{25}, \integer{50}, \integer{75}, and \integer{90}-th percentiles per checkpoint.}
    \label{fig:cka_main_plot}
\end{figure}
Scaling the number of parameters in language models (LMs) has provided impressive performance gains on a variety of tasks \citep{hendrycks2020measuring} and has become the \emph{de facto} standard to make progress in model design \citep[e.g.,][]{chowdhery2023palm}.
Small LMs, however, remain essential as they are more practical: lower training and inference costs result in a smaller environmental impact \citep{schwartz2020greenai}. 
Small LMs empower individuals to train on proprietary data by requiring fewer resources, enhancing data privacy \citep{huang2022large} and democratising access to language modelling technology \citep{bender2021dangers}.
However, for the same data and computational budget, small LMs (unsurprisingly) underperform larger ones \citep{biderman-etal-2023-pythia} and (importantly) their performance tends to degrade in the late pretraining phase, a phenomenon termed \textit{saturation} by \citet{godey2024small} \footnote{Subsequent references in this paper to \textit{saturation} align with the concept introduced by \citet{godey2024small}.}. Saturation is typically attributed to the \enquote{limited representational capacity} of small LMs; besides this anectodal justification, our understanding of its causes is still limited. 

Recently, \citet{godey2024small} linked saturation to the reduced variability of the output embeddings of LMs caused by the mismatch between the hidden model dimension and the vocabulary size \citep{yang2018breaking}. Specifically, the last layer of LMs maps the hidden representation of random tokens to output embeddings with high cosine similarity.\footnote{This issue is termed \emph{anisotropy} \citep{ethayarajh-2019-contextual}.}

In this paper, we use the Pythia model suite \citep{biderman-etal-2023-pythia} to provide orthogonal analyses that consider models' training dynamics.  
First, we study how the activations of the $\attention$ and $\mlp$ layers converge to their final state across LMs of different sizes. Then, we relate the difference in convergence behaviour across sizes to the effective rank of their parameters: layers whose activations converge later in training span a smaller fraction of their dimensions.

Specifically, we first use the \textbf{Centered Kernel Alignment} \citep[$\cka$;][]{kornblith2019similarity} metric to measure the similarity of layers' activations across checkpoints. We observe that larger LMs converge faster and more smoothly to their final state. As shown in \cref{fig:cka_main_plot}, within the first \q{20}{\percent} of training nearly all layers in the larger LM (\twobil) resemble their final state, while most layers in the smaller LM (\sixmil) remain different for most of training.

We then find a strong correlation between the convergence pattern of a layer's activations and the rank of its parameters and gradients. We introduce the concept of \defn{proportional effective rank} (\cref{sec:methodology}) to consistently compare these effective ranks 
across model sizes. Our analyses highlight training inefficiencies in small-scale LMs, paving the way for targeted improvements in future work.

\section{Related Work}\label{sec:related_work}

Prior work has studied various learning dynamics of the Pythia suite, including memorisation \citep{biderman-etal-2023-pythia, lesci-etal-2024-causal}, training data influence \citep{liu2024training}, and statistics of learned embeddings \citep{belrose2024neural}.
Related to our work, \citet{godey2024small} examine the differences in the rank of the unembedding matrix (mapping from hidden representations to tokens) across model sizes, known as the softmax bottleneck \citep{yang2018breaking}. Unlike their findings, we focus on the convergence dynamics of all layers.

Similarity metrics like $\cka$ and Singular Vector Canonical Correlation Analysis (SVCCA) are widely used to analyse language model properties. \citet{nguyen2020wide} find that architectural decisions, such as model width and depth, affect hidden representation similarity. \citet{wu2020similarity} show that models within the same architectural family share similar hidden structures, a similarity that persists even in fine-tuned models \citep{phang2021fine}. Additionally, SVCCA has been used to study token representation distribution in multilingual models \citep{singh2019bert} and syntactic element learning in monolingual models \citep{saphra2019understanding}.
Most similar to our work, \citet{brown2023understanding} use representation similarity metrics, including $\cka$, to study Pythia generalisation capabilities. However, our study is the first to use the $\cka$ metric to examine the convergence dynamics of layers' activations across model sizes.

\section{Methodology}\label{sec:methodology}

We first describe the residual stream view of transformer-based models and define layers' activations. Then, we introduce the $\cka$ and proportional effective rank metrics.

\paragraph{The \textit{Residual Stream} view.}
The residual stream view of the transformer architecture \citep{vaswani-etal-2017-attention} is an analytical framework to study how information flows through its layers \citep{elhage-etal-2021-mathematical}. 
This conceptualisation focuses on the residual connections as they provide a direct reference to the inputs. Specifically, the set of residual connections across layers is termed the \defn{residual stream}. Each layer can be seen as providing modifications to the residual stream via addition operations.
Layers have two main components, $\attention$ and $\mlp$, that sequentially update the residual stream.
Formally, a sequence of $\seqlen$ tokens $\sequence = \langle \token_1, \smalldots, \token_\seqlen\rangle$ is first converted into a matrix $\xdata_0 \mathop{\in} \R^{\mathop{\seqlen\times\residualdim}}$ by the embedding layer: each column is a token representation of size $\residualdim$. Then, each layer $\layer\mathop{\in}\{1,\smalldots, \numlayers\}$ updates these representations as follows:
\begin{align}\label{eq:residual_stream}
    \xdata' &= \xdata_{\layer-1}  + \dashuline{\attention(\xdata_{\layer-1})} \\
    \xdata_\layer &= \xdata' + \dashuline{\mlp(\xdata')}
\end{align}
Finally, the $\seqlen$-th column of $\xdata_\numlayers$ is used to predict the $(\seqlen\mathop{+}1)$-th token. More details in \cref{app:residual-stream}.

\paragraph{\textit{Activations} and \textit{Parameters}.}
The updates to the residual stream---\dashuline{underlined} in \cref{eq:residual_stream}---are the layer's \defn{activations} and have the same dimensions as the residual stream, i.e., $\R^{\mathop{\seqlen\times\residualdim}}$. Both $\attention$ and $\mlp$ first project, or \enquote{read}, the residual stream into lower-dimensional intermediate representations; then project these representations back, or \enquote{write}, into the residual stream. Here, we study the behaviour of the \defn{parameters} that write to the residual stream.
We use $\actatt$ and $\actmlp$ to denote the activations and $\attweight$ and $\mlpweight$ to denote the parameters of, respectively, $\attention$ and $\mlp$.

\paragraph{Activations' Similarity.}
Given a set of activations, either $\actatt$ or $\actmlp$, of a layer $\layer$ at a particular checkpoint $\checkpoint$, $\act_{\layer, \checkpoint}$, we measure how similar they are to those at the last checkpoint $\numcheckpoints$, $\act_{\layer, \numcheckpoints}$, using the linear variant of the Centered Kernel Alignment metric \citep[$\cka$;][]{kornblith2019similarity}:
\begin{align}\label{eq:cka}
    \cka(\actcenter_{\checkpoint}, \actcenter_{\numcheckpoints}) = \frac{\norm{\actcenter_{\checkpoint}{}^{\top}\, \actcenter_{\numcheckpoints}}^2_F}{\norm{\actcenter_{\checkpoint}{}^{\top}\,\actcenter_{\checkpoint}}_F\; \;\norm{\actcenter_{\numcheckpoints}{}^{\top}\,\actcenter_{\numcheckpoints}}_F}
\end{align}
where $\actcenter$ denotes the centred activations, and $\norm{\cdot}_F$ is the Frobenius norm; we omit the layer subscript $\layer$ for clarity.
We compute \cref{eq:cka} for both $\actatt$ and $\actmlp$ across all layers and checkpoints throughout training, allowing us to examine the convergence dynamics of each layer's activations.

\paragraph{Parameters' \textit{Proportional Effective Rank}.}
Let $\hiddendim$ be the dimension of the intermediate representation of either $\attention$ or $\mlp$. For a layer $\layer$, let $\vtheta_{\layer} \in \R^{\mathop{\residualdim\times\hiddendim}}$ be the subset of parameters of either $\attweight$ or $\mlpweight$ that comprise the matrix that projects from the hidden space into the residual stream.
We measure the effective number of dimensions onto which $\vtheta_{\layer}$ projects the intermediate representations using the definition of  \defn{effective rank} introduced in \citet{roy-vetterli-2007-effective}. The effective rank is computed as the entropy over the normalised singular values of the parameter matrix $\vtheta_{\layer}$, that is:
\begin{align}\label{eq:per}
    \er(\vtheta_{\layer}) = \exp \left( -\sum_{k=1}^K \frac{\sigma_k}{\norm{\sigma}_1} \; \log \frac{\sigma_k}{\norm{\sigma}_1}\right)
\end{align}
where $\sigma = \langle\sigma_1, \smalldots, \sigma_K\rangle$ is the vector of singular values and $\norm{\cdot}_1$ is the $\ell_1$ norm.
In this paper, we introduce the notion of a \defn{proportional effective rank} ($\per$) computed as the effective rank normalised by the number of hidden dimensions: 
\begin{align}
    \per(\vtheta_{\layer}) = \er(\vtheta_{\layer}) \, / \, \hiddendim
\end{align}
The $\per$ allows us to compare the effective rank of layers with different sizes consistently. We compute the $\per$ of both $\attweight$ and $\mlpweight$, as well as the gradients of these parameters, across all layers and checkpoints throughout training. 

\begin{figure*}[h!]
    \centering
    \includegraphics[width=\linewidth]{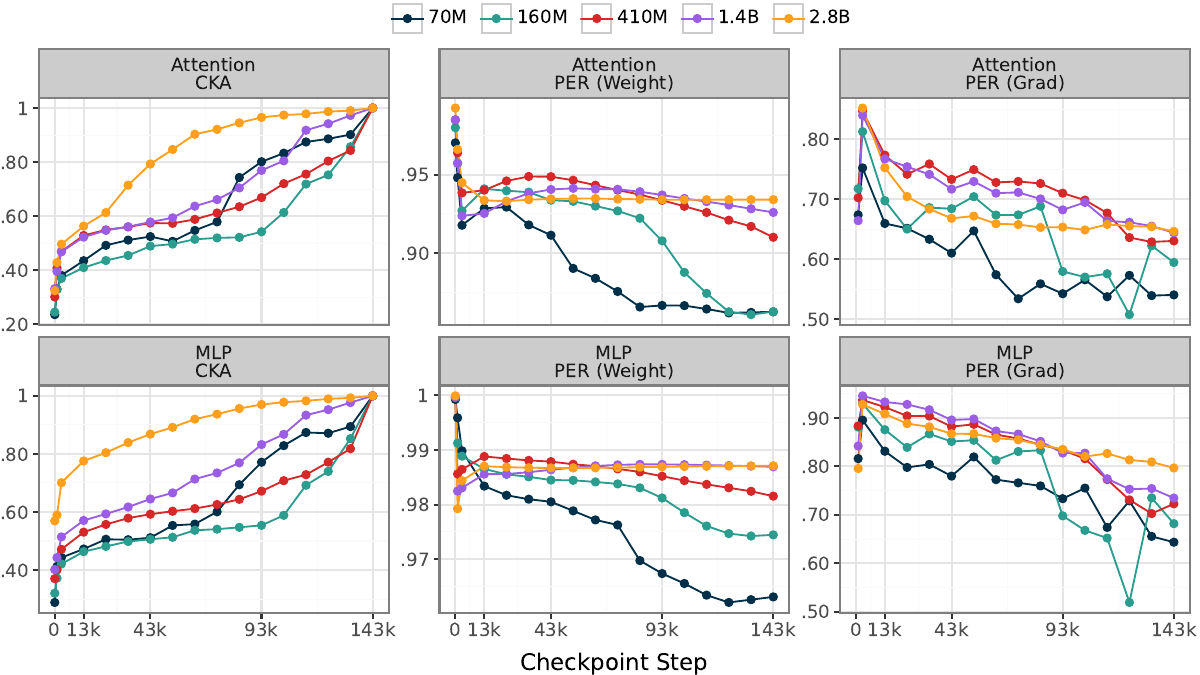}
    \vspace{-15pt}
    \caption{$\cka$ similarity (current vs.\ last checkpoint) of layers' activations (first column), $\per$ of layers' parameters (second column) and gradients (third column) for $\attention$ (top row) and $\mlp$ (bottom row) in Pythia \sevenmil, \sixmil, \fourmil, \onebil, and \twobil averaged (mean) across layers per each checkpoint.}
    \label{fig:main-results}
\end{figure*}

\section{Experimental Setup}\label{sec:experimental_setup}

We use the Pythia model suite \citep{biderman-etal-2023-pythia}, composed of \integer{8} transformers of different sizes trained for \q{143}{\thousand} steps on the deduplicated\footnote{There exists a non-deduplicated (or standard) version of the Pile dataset used to train a first version of the Pythia suite.} version of the Pile dataset \citep{gao-etal-2020-pile, biderman-etal-2022-datasheet}.
Intermediate checkpoints are available every \q{1}{\thousand} steps and at log-spaced intervals early in training.
To comply with our computational budget, we consider models up to \q{2.8}{\billion} parameters---i.e., \sevenmil, \sixmil, \fourmil, \onebil, and \twobil---evaluated at the following steps: \integer{0}, all log-spaced steps $\{1, 2, 4, \smalldots, 512\}$, \q{1}{\thousand}, \q{3}{\thousand}, and then every \q{10}{\thousand} steps up to \q{143}{\thousand}.
We evaluate each checkpoint on the last batch of the training set and collect its activations.
More details in \cref{app:implementation_details}.

\section{Results}\label{sec:results}

Our analyses reveal quantitative differences in the learning dynamics of layers across model sizes.

\begin{result}[Activations of larger models converge faster and more monotonically to their final state than those of smaller models]
\label{result:cka}
As observed in \cref{fig:main-results} (first column), larger models show, on average, earlier convergence of $\attention$ and $\mlp$ activations. For example, by \q{20}{\percent} of training, the $\cka$ score in \twobil is \float[1]{0.8} for $\mlp$ and \float[1]{0.7} for $\attention$, where in \sevenmil and \sixmil it is around \float[1]{.5}.
This fast convergence pattern holds across layers, as shown by the distributions in \cref{fig:cka_main_plot}.
\end{result}

\begin{result}[Activations of earlier layers converge faster, regardless of the model size]
Across model sizes, earlier layers' activations converge faster to their final state than those of later layers. As shown in \cref{fig:cka-layer-wise-lines} (\cref{app:layerwise-convergence-figures}), the faster average convergence in larger models is due to more of their later layers converging earlier, whereas smaller models' layers only reach their final state towards the end of training.
\end{result}

\vspace{9pt}
Based on recent work that identifies parameter rank differences across model sizes \citep{godey2024small}, in the next paragraphs, we study whether the different convergence behaviours are related to the effective rank of layers' parameters and gradients.
\vspace{9pt}

\begin{result}[Parameters of layers in larger models proportionally span more dimensions] 
\label{result:weight-effective-rank} 
Parameters in layers of larger models span a slightly larger fraction of their available dimensions compared to smaller models, as shown in \cref{fig:main-results} (second column). 
Moreover, the $\per$ of larger models stabilises early, while it keeps decreasing throughout training for smaller ones. This finding is further underscored when visualising the $\per$ for each layer, as shown in \cref{fig:per_weight-layer-wise-lines} (\cref{app:layerwise-per_weight-figures}); we observe that in smaller models the $\per$ of later layers tends to decrease over the course of training, while in larger models the $\per$ of all layers stabilises early in training. This difference is even more pronounced in the $\per$ of these layers' gradients, as shown in \cref{fig:main-results} (third column).
\end{result}

\begin{result}[Parameters of layers in larger models receive gradient updates along proportionally more dimensions]
\label{result}
The $\per$ of gradients reflects the proportion of the learning signal transmitted by the gradients relative to the available parameter dimensions. In \cref{fig:main-results} (third column), we observe that throughout training gradients in larger models consistently span a larger fraction of the available dimensions, with this fraction gradually decreasing over time. In contrast, smaller models display more variability. At first glance, the averaged $\per$ of gradients in the $\attention$ layer of the \twobil model might appear to contradict the observed trend. However, this discrepancy is clarified when examining the $\per$ of gradients across individual layers, as shown in \cref{fig:per_grad-layer-wise-lines} (\cref{app:layerwise-per_grad-figures}). Once again, we observe that the $\per$ of gradients in later layers of smaller models are less stable compared to larger models. The reason the average $\per$ of gradients in the $\attention$ layer of the \twobil model is smaller than in smaller models is that, early in training, all layers of the larger model stabilise at their final values. At this stage, the stabilised layers of the larger model have lower gradient $\per$ values compared to those of smaller models, which have not yet converged. Overall, our findings suggest that layers in larger models converge both more quickly and tend to receive proportionally larger rank updates during training.
\end{result}

\begin{result}[The dynamics of the parameters' effective rank and the activations' convergence patterns are correlated]
We investigate the correlation between a layer’s activations convergence rate and the rank of its parameters and gradients. Broadly, we find that layers with higher effective rank in both weights and gradients converge faster.
To measure this correlation, we first create two binary variables for each layer indicating whether (i) it converges early in training and (ii) maintains a stable $\per$ throughout training. Then, we calculate the Matthew's Correlation Coefficient between these two statistics across layers and report them in \cref{tab:model_correlation}.
Specifically, for each layer of a given model, we determine whether that layer exhibits early activations' convergence and large and stable parameters' and gradients' $\per$s (relative to other model layers) using the following heuristics: 
\begin{itemize}
    \item \textbf{Early activations' convergence.} Activations' $\cka \mathop{\geq} \float[2]{0.45}$ by the first \q{10}{\percent} of training (applies to both the $\attention$ and $\mlp$ layers).
    
    \item \textbf{Large parameters' $\per$.} Parameters' $\per \mathop{\geq} \float[2]{0.95}$ by the end of training (applies to both the $\attention$ and $\mlp$ layers).
    
    \item \textbf{Large gradients' $\per$.} We note that gradients' $\per$ slightly decreases throughout training for each model size. Rather than choosing a fixed value to determine large and stable gradients' $\per$s, we dynamically set the threshold at \q{90}{\percent} of the largest $\per$ attained by any layer at the end of training.
\end{itemize}
We observe a strong correlation for the $\attention$ layers across model sizes. For the $\mlp$ layers, the correlation with the gradients' $\per$ is strong for models up to \onebil, while the correlation with the parameters' $\per$ is strong only for the \sevenmil model. We hypothesise that this discrepancy can be explained by the fact that $\mlp$ layers have a large $\per$ throughout training across all model sizes, apart from those of the \sevenmil model. 

While these results are correlational, they provide a foundation for future work to test whether methods that specifically increase the PER of layers' parameters and gradients induce faster convergence of the layers' activations in small models.
\end{result}

\begin{table}[!t]
    \centering
    \begin{tabular}{crrrr}
\toprule
 \textbf{Size} & $\attweight$  & $\nabla \attweight$ & $\mlpweight$ & $\nabla\mlpweight$\\ 
\midrule
\sevenmil  & \float[2]{1.000} & \float[2]{1.000} & \float[2]{0.632} & \float[2]{1.00} \\ 
\sixmil & \float[2]{1.000} & \float[2]{0.845} & \float[2]{0.357} & \float[2]{0.714} \\ 
\fourmil & \float[2]{0.837} & \float[2]{0.916} & \float[2]{0.192} & \float[2]{0.777} \\ 
\onebil & \float[2]{0.775} & \float[2]{0.845} & \float[2]{0.209} & \float[2]{0.641} \\ 
\twobil & \float[2]{0.728} & \float[2]{0.521} & \float[2]{0.112} & \float[2]{0.179} \\ 
\bottomrule
\end{tabular}
    \caption{Matthew's Correlation Coefficient 
    between binary variables indicating whether a given layer converges early in training and whether it maintains a stable PER of the parameters ($\vtheta$) and gradients ($\nabla\vtheta$) throughout training for both $\attention$ and $\mlp$.}
    \label{tab:model_correlation}
    \vspace{-8pt}
\end{table}

\vspace{-6pt}
\section{Conclusion}\label{sec:conclusion}

Our study highlights disparities in the learning dynamics of small and large LMs. Using the Pythia model suite, we demonstrate that layers' activations in larger models converge faster and more monotonically to their final state.
We correlate this phenomenon with the larger $\per$ in the parameters and gradients of larger models. 
Our analyses expand our understanding of training inefficiencies in small models and provide insights for future work to address them, e.g., by developing methods that increase the $\per$ of layers’ parameters.

\section*{Ethical Impact}\label{sec:ethical-impact}

Our work is part of a greater effort in Green AI \cite{schwartz2020greenai} to lower the environmental footprint of training and using language models. We acknowledge, however, that small language models are prone to the same types of biases as large language models that are encoded through the data the models are trained on; the Pile is known to contain gender and racial biases \cite{gao-etal-2020-pile}.

\section*{Limitations}\label{sec:limitations}

We experiment only with the Pythia model suite and the Pile dataset. It is unclear to what extent our findings translate to other models and datasets (including datasets in languages other than English). Moreover, because of our restricted computational budget, we are limited in our ability to thoroughly study larger language models. 
The largest models we experiment with are still relatively small given the scale of currently available open-source large language models (in the hundreds of billions). Finally, the relationship we find between the $\cka$ similarity scores and the proportional effective rank is purely correlational: in future work, we aim to use our results to guide targeted interventions to assess whether the relationship we found is causal, i.e. whether increasing the effective rank of a layer can increase its convergence speed.

\section*{Acknowledgements}

We thank the anonymous reviewers for their helpful comments, which helped us improve the paper.

The experiments reported in this paper were performed using resources provided by the Cambridge Service for Data Driven Discovery (CSD3) operated by the University of Cambridge Research Computing Service, provided by Dell EMC and Intel using Tier-2 funding from the Engineering and Physical Sciences Research Council (capital grant EP/T022159/1), and DiRAC funding from the Science and Technology Facilities Council. Richard Diehl Martinez is supported by the Gates Cambridge Trust (grant OPP1144 from the Bill \& Melinda Gates Foundation).

\setlength{\intextsep}{0pt}
\setlength{\columnsep}{8pt}
\begin{wrapfigure}{l}{0.45\columnwidth}
    \includegraphics[width=0.45\columnwidth]{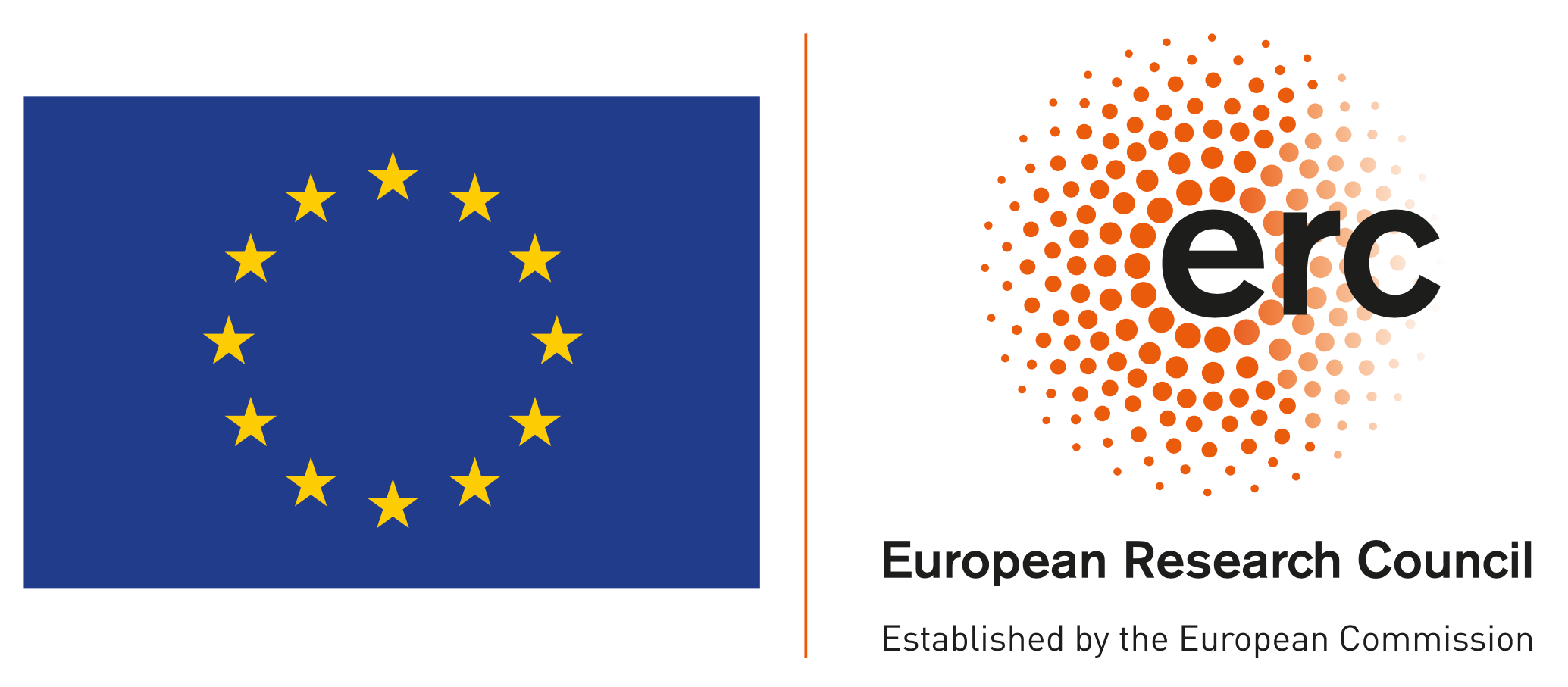}
\end{wrapfigure}
\noindent Pietro received funding from the European Research Council (ERC) under the European Union’s Horizon 2020 Research and Innovation programme grant AVeriTeC (Grant agreement No. 865958).

\bibliography{main.bib}

\clearpage
\appendix
\onecolumn

\section{The Residual Stream View} \label{app:residual-stream}

The residual stream is a mathematical formalization through which to study how transformer models process inputs \citep{elhage-etal-2021-mathematical}. Under this framework, each of the $L$ layers of a transformer model processes a series of input tokens $\sequence = \langle \token_1, \smalldots, \token_\seqlen\rangle$ consecutively and communicate the result of their computation for each token to subsequent layers via a residual stream of dimension $\residualdim$. 
The reading, processing, and writing of the residual stream occur independently in each $\attention$ head via combinations of the query, key, value and output matrices, $W_Q$, $W_K$, $W_V$, $W_O$: The \textbf{query-key circuit}, $W_Q^{\top}W_K$, of the $\attention$ mechanism controls how the residual stream should be recomposed, and the \textbf{output circuit}, $W_OW_V$, writes to the residual stream an update that is mediated by the query-key circuit. The write operation of each $\attention$ head is of low rank relative to $\residualdim$. After each $\attention$ head has written to the residual stream, a bottleneck $\mlp$ projection performs a full-rank transformation on the residual stream.  Due to their pivotal role in updating the state of the residual stream, our work analyses the learning dynamics of the two operations that write to the residual stream: the output circuit of each head of the $\attention$ layer---that we refer to as $\attention$---and the $\mlp$ projection layer---that we denote $\mlp$ for conciseness.

\section{Implementation Details}\label{app:implementation_details}

We implement all experiments using the \myemph{PyTorch} framework \citep{paszke-etal-2019-pytorch}. We access the Pythia models through the \myemph{transformers} library \citep{wolf-etal-2020-transformers}.

\subsection{Hardware Details}

We use a server with one \myemph{NVIDIA A100 80GB PCIe}, \integer{32} CPUs, and \integer{32} GB of RAM for all experiments. Collecting model activations for all analyses required in total about 24 GPU hours. Below, we report a subset of the output of the \myemph{lscpu} command:

\begin{tcolorbox}[left=5pt,right=5pt,top=5pt,bottom=5pt]
\small
\begin{verbatim}
Architecture:        x86_64
CPU op-mode(s):      32-bit, 64-bit
Address sizes:       46 bits physical, 
                     48 bits virtual
Byte Order:          Little Endian
CPU(s):              32
On-line CPU(s) list: 0-31
Vendor ID:           GenuineIntel
Model name:          Intel(R) Xeon(R)
                     Silver 4210R CPU
                     @ 2.40GHz
CPU family:          6
Model:               85
Thread(s) per core:  1
Core(s) per socket:  1
Socket(s):           8
Stepping:            7
BogoMIPS:            4800.11
\end{verbatim}
\end{tcolorbox}

\subsection{The Pythia Suite}
\label{subsection:pythia-details}

We use the publicly available Pythia model suite \citep{biderman-etal-2023-pythia}, which was trained on the Pile \citep{gao-etal-2020-pile, biderman-etal-2022-datasheet}. Both the preprocessed training data and intermediate checkpoints are publicly available.\footnote{\href{https://github.com/EleutherAI/pythia}{\myemph{github.com/EleutherAI/pythia}} (Apache License 2.0).} 

\paragraph{Data.}
The Pile is a \q{300}{\billion}-token curated open-source collection of English documents, spanning a wide range of domains (e.g. books, academic publications, Wikipedia). \footnote{\href{https://github.com/EleutherAI/the-pile}{\myemph{github.com/EleutherAI/the-pile}} (MIT License).}  
The deduplicated version of the dataset is obtained by applying a near-deduplication method based on \myemph{MinHashLSH} and has \q{207}{\billion} tokens.
Thus, models trained on this version of the dataset are trained for circa \float[1]{1.5} epochs to keep an equal token count relative to the non-deduplicated versions.
The dataset is shuffled, tokenised, and \enquote{packed} into sequences of \integer{2049} tokens with no end-of-document token\footnote{\href{https://github.com/EleutherAI/pythia/issues/123}{\myemph{github.com/EleutherAI/pythia/issues/123}}.}.
Noticeably, the packing process implies that the second half-epoch of deduplicated data contains the same documents but not necessarily the same sequences. 
By design, each sequence can pack multiple documents and tokens can attend across document boundaries.

\paragraph{Models.}
The Pythia model suite is composed of 16 models: transformers of \integer{8} different sizes trained on the Pile as-is and deduplicated.
All model sizes were trained using a cosine learning rate schedule with warm-up, the same data order, and a batch size of \integer{1024} sequences, resulting in exactly \q{143}{\thousand} optimization steps.
Checkpoints are available at initialization (step \integer{0}), and after every \q{1}{\thousand} iterations (steps \q{1}{\thousand}-\q{143}{\thousand}) resulting in \integer{144} checkpoints evenly spaced throughout training. 
Additionally, log-spaced checkpoints are available early in training (steps $\{2^i\}_{i=0}^{9}$). In \cref{tab:model_hparams} we report more details about the architecture and training hyper-parameters of the models in the suite.

\begin{table}[!t]
    \centering
    \begin{tabular}{lccccccc}
\toprule
\textbf{Size} & $\numlayers$ & $\residualdim$ & \textbf{\# Heads} & \textbf{Head Dim.} & \textbf{Batch Size} & \textbf{Learning Rate} & \textbf{Checkpoints} \\
\midrule
\sevenmil & \integer{6} & \integer{512} & \integer{8} & \integer{64} & \q{2}{\million} & \snum{1e-3} & \href{https://huggingface.co/EleutherAI/pythia-70m}{\myemph{Standard}}, \href{https://huggingface.co/EleutherAI/pythia-70m-deduped}{\myemph{Deduped}} \\
\sixmil & \integer{12} & \integer{768} & \integer{12} & \integer{64} & \q{2}{\million} & \snum{6e-4} & \href{https://huggingface.co/EleutherAI/pythia-160m}{\myemph{Standard}}, \href{https://huggingface.co/EleutherAI/pythia-160m-deduped}{\myemph{Deduped}} \\
\fourmil & \integer{24} & \integer{1024} & \integer{16} & \integer{64} & \q{2}{\million} & \snum{3e-4} & \href{https://huggingface.co/EleutherAI/pythia-410m}{\myemph{Standard}}, \href{https://huggingface.co/EleutherAI/pythia-410m-deduped}{\myemph{Deduped}} \\
\onebil & \integer{24} & \integer{2048} & \integer{16} & \integer{128} & \q{2}{\million} & \snum{2e-4} & \href{https://huggingface.co/EleutherAI/pythia-1.4b}{\myemph{Standard}}, \href{https://huggingface.co/EleutherAI/pythia-1.4b-deduped}{\myemph{Deduped}} \\
\twobil & \integer{32} & \integer{2560} & \integer{32} & \integer{80} & \q{2}{\million} & \snum{1.6e-4} & \href{https://huggingface.co/EleutherAI/pythia-2.8b}{\myemph{Standard}}, \href{https://huggingface.co/EleutherAI/pythia-2.8b-deduped}{\myemph{Deduped}} \\
\bottomrule
\end{tabular}

    \caption{Details on the architecture and training hyper-parameters for models in the Pythia suite used in this paper.}
    \label{tab:model_hparams}
\end{table}

\clearpage

\section{Layer-wise CKA Convergence Dynamics}
\label{app:layerwise-convergence-figures}

\begin{figure*}[h!]
    \centering
    \includegraphics[width=0.90\linewidth]{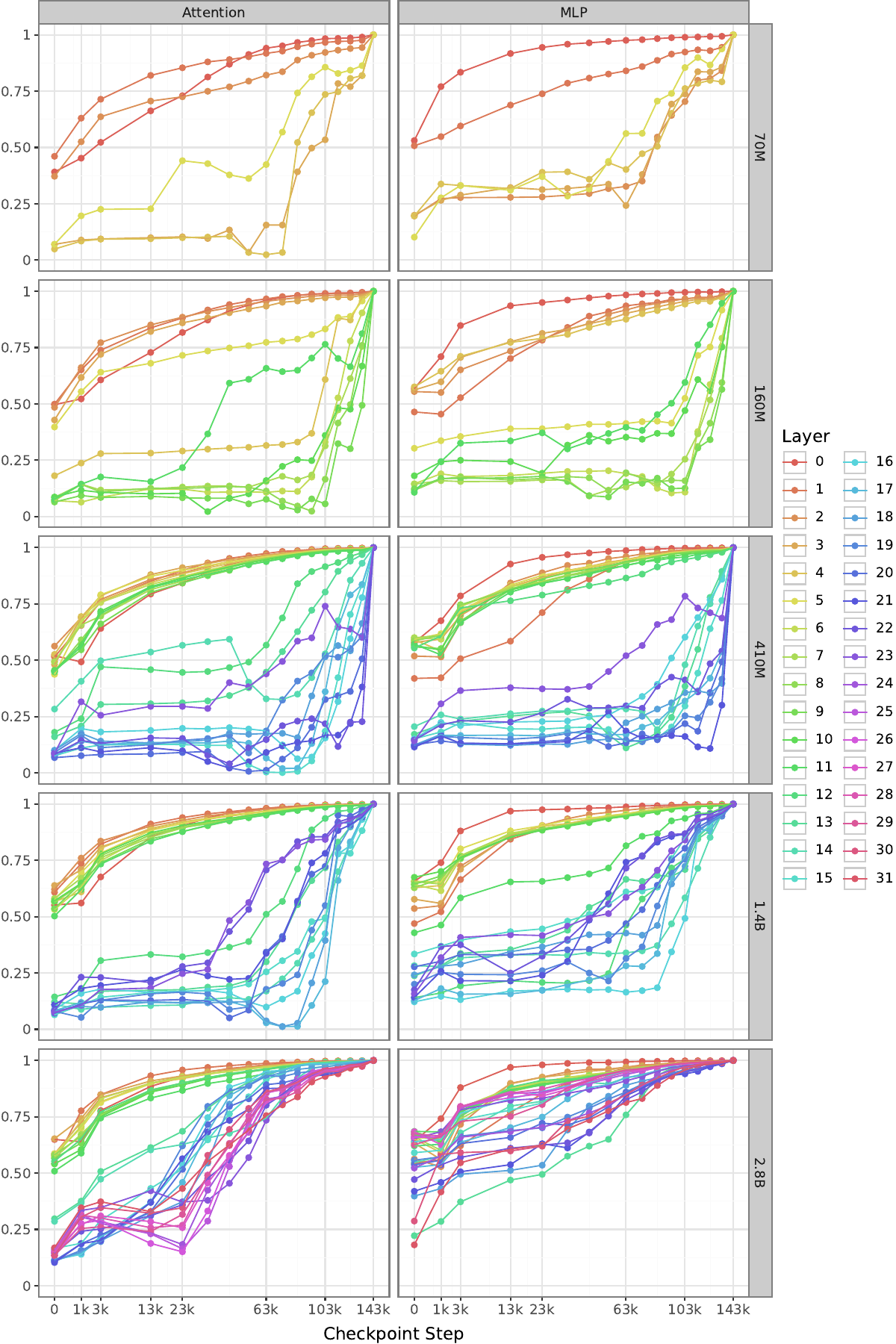}
    \vspace{-5pt}
    \caption{$\cka$ similarity (current vs last checkpoint) of the activations of $\attention$ and $\mlp$ in each layer of Pythia \q{70}{\million}, \q{160}{\million}, \q{410}{\million}, \q{1.4}{\billion} and \q{2.8}{\billion} throughout training.}%
    \label{fig:cka-layer-wise-lines}
\end{figure*}

In \cref{fig:cka-layer-wise-lines}, we visualise the activations' $\cka$ convergence dynamics of layers in different models as a colour-coded line plot.
\vspace{0.2cm}

\section{Layer-wise PER Weight Dynamics}
\label{app:layerwise-per_weight-figures}

\begin{figure*}[h!]
    \centering
    \includegraphics[width=0.90\linewidth]{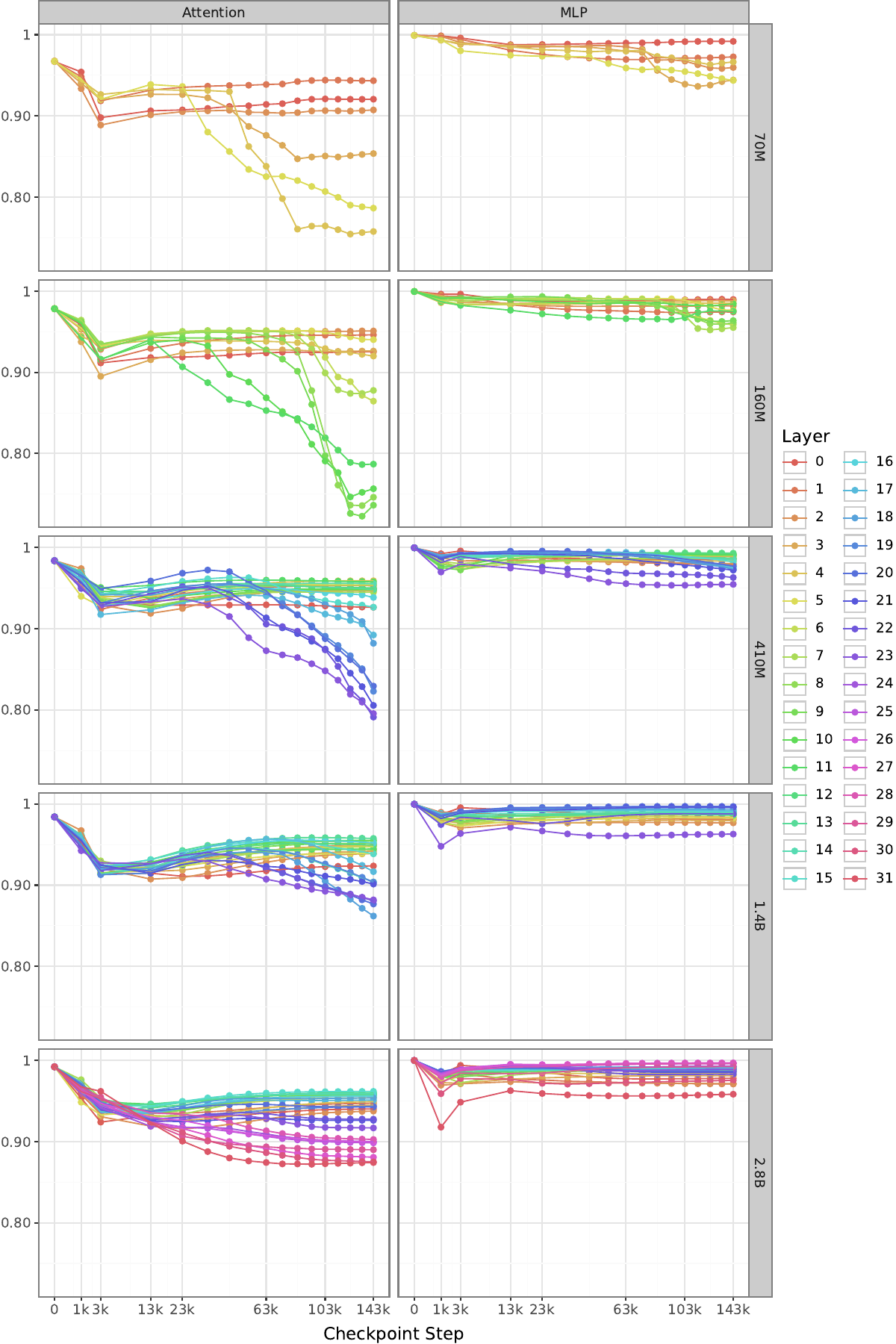}
    \vspace{-5pt}
    \caption{$\per$ of the weight matrices of $\attention$ and $\mlp$ in each layer of Pythia \q{70}{\million}, \q{160}{\million}, \q{410}{\million}, \q{1.4}{\billion} and \q{2.8}{\billion} throughout training.}%
    \label{fig:per_weight-layer-wise-lines}
\end{figure*}

In \cref{fig:per_weight-layer-wise-lines}, we visualise the learning dynamics of the $\per$ of weight matrices of layers in different models as a colour-coded line plot. 
\vspace{0.2cm}

\section{Layer-wise PER Gradient Dynamics}
\label{app:layerwise-per_grad-figures}

\begin{figure*}[h!]
    \centering
    \includegraphics[width=0.90\linewidth]{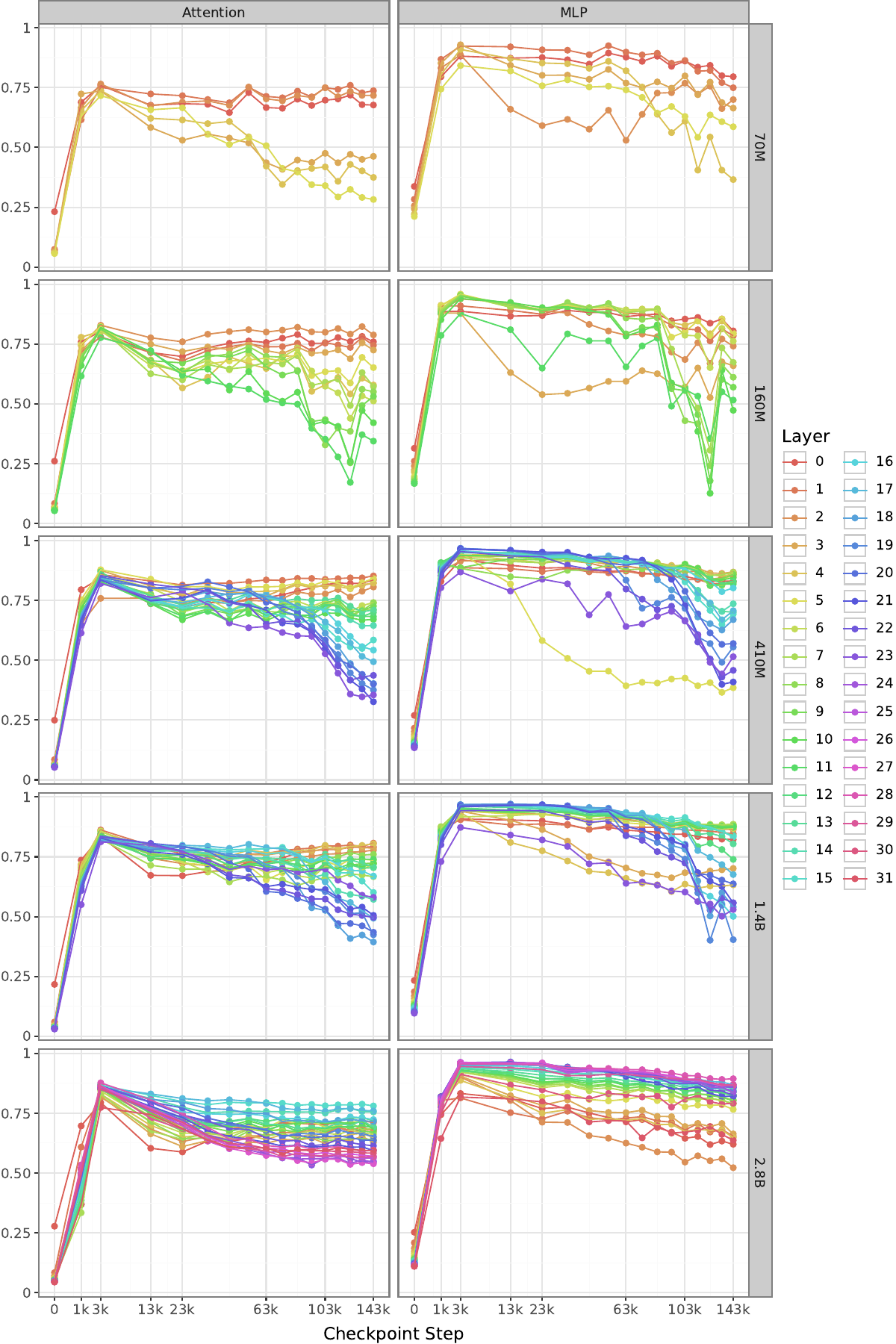}
    \vspace{-5pt}
    \caption{$\per$ of the gradients of the weight matrices of $\attention$ and $\mlp$ in each layer of Pythia \q{70}{\million}, \q{160}{\million}, \q{410}{\million}, \q{1.4}{\billion} and \q{2.8}{\billion} throughout training.}%
    \label{fig:per_grad-layer-wise-lines}
\end{figure*}

In \cref{fig:per_grad-layer-wise-lines}, we visualise the learning dynamics of the $\per$ of gradients of layers in different models as a colour-coded line plot. 
\vspace{0.2cm}

\end{document}